\documentclass[10pt,twocolumn,letterpaper]{article}

\usepackage{cvpr}
\usepackage{times}
\usepackage{epsfig}
\usepackage{graphicx}
\usepackage{amsmath}
\usepackage{amssymb}
\usepackage{color}
\usepackage{xcolor}
\usepackage{breqn}
\usepackage{booktabs}       
\usepackage{chngpage}

\definecolor{darkpastelgreen}{rgb}{0.01, 0.75, 0.24}

\usepackage[pagebackref=true,breaklinks=true,letterpaper=true,colorlinks,bookmarks=false]{hyperref}

\cvprfinalcopy 

\def\cvprPaperID{1365} 

\ifcvprfinal\fi
\begin{document}

\title{Dense Depth Posterior (DDP) from Single Image and Sparse Range}

\author{Yanchao Yang, \ \ \ Alex Wong, \ \ \ Stefano Soatto\\
UCLA Vision Lab\\
University of California, Los Angeles, CA 90095\\
{\tt\small \{yanchao.yang, alexw, soatto\}@cs.ucla.edu }
}

\maketitle
\thispagestyle{empty}

\begin{abstract}
We present a deep learning system to infer the posterior distribution of a dense depth map associated with an image, by exploiting sparse range measurements, for instance from a lidar. While the lidar may provide a depth value for a small percentage of the pixels, we exploit regularities reflected in the training set to complete the map so as to have a probability over depth for each pixel in the image. We exploit a Conditional Prior Network, that allows associating a probability to each depth value given an image, and combine it with a likelihood term that uses the sparse measurements. Optionally we can also exploit the availability of stereo during training, but in any case only require a single image and a sparse point cloud at run-time. We test our approach on both unsupervised and supervised depth completion using the KITTI benchmark, and improve the state-of-the-art in both. Code is available at: \url{https://github.com/YanchaoYang/Dense-Depth-Posterior}
\end{abstract}

\vspace{-0.4em}
\section{Introduction}
\vspace{-0.4em}
There are many dense depth maps that are compatible with a given image and a sparse point cloud. Any point-estimate, therefore, depends critically on the prior assumptions made. Ideally, one would compute the entire posterior distribution of depth maps, rather than a point-estimate. The posterior affords to reason about confidence, integrating evidence over time, and in general, is a (Bayesian) sufficient representation that accounts for all the information in the data.

\begin{figure}[ht]
  \centering
  \includegraphics[width=0.45\textwidth]{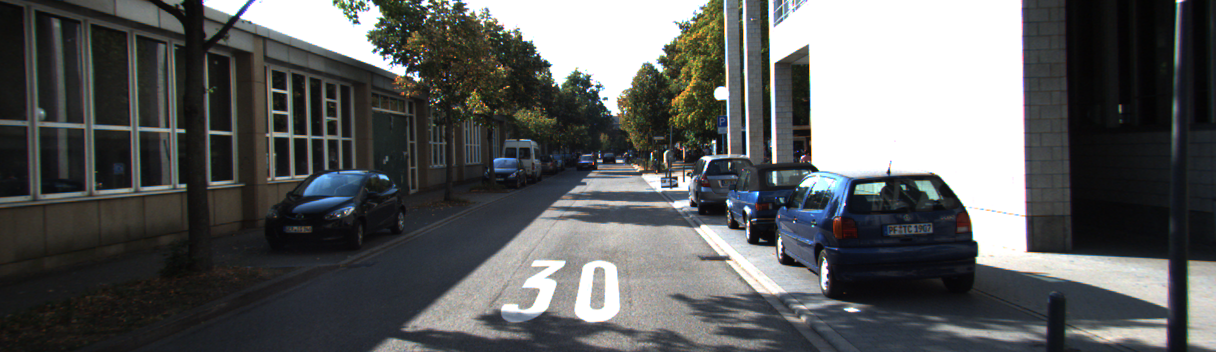}
  \includegraphics[width=0.45\textwidth]{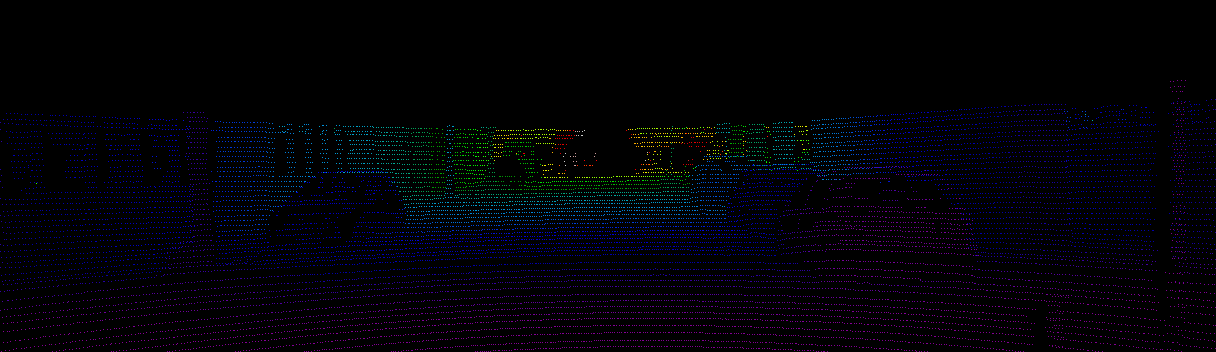}
  \includegraphics[width=0.45\textwidth]{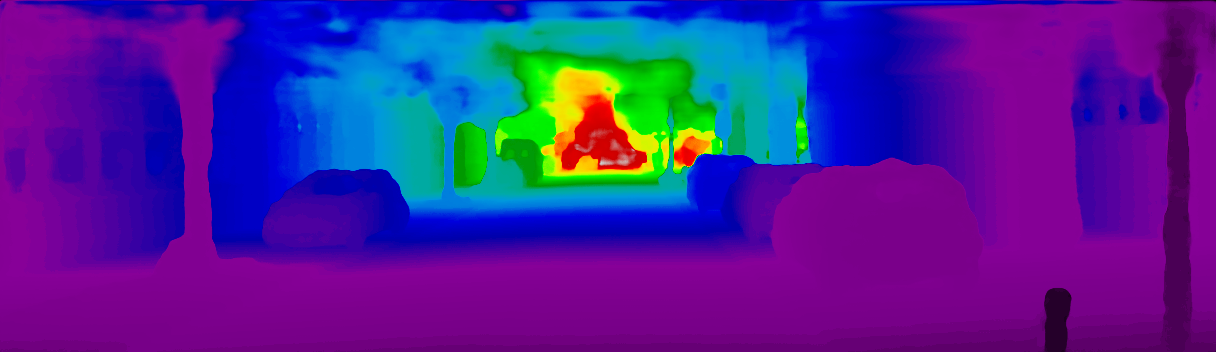}
  \includegraphics[width=0.45\textwidth]{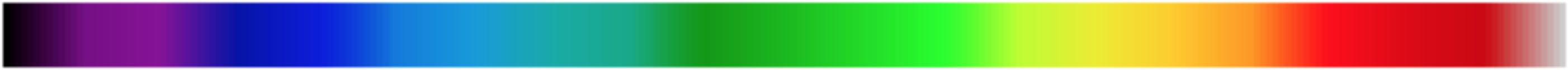}
  \setlength{\belowcaptionskip}{-15pt}
  \caption{An image (top) is insufficient to determine the geometry of the scene; a point cloud alone (middle) is similarly ambiguous. Lidar returns are shown as colored points, but black regions are uninformative: Are the black regions holes in the road surface, or due to radiometric absorption? Combining a single image, the lidar point cloud, and previously seen scenes allows inferring a dense depth map (bottom) with high confidence. Color bar from left to right: zero to infinity.}
  \label{fig:task-illustration}
\end{figure}

{\bf Motivating application.} In autonomous navigation, a sparse point cloud from lidar may be insufficient to make planning decisions: Is the surface of the road in Fig.~\ref{fig:task-illustration} (middle, better viewed when enlarged) littered with pot-holes, or is it a smooth surface? Points that are nearby in image topology, projecting onto adjacent pixels, may be arbitrarily far in the scene. For instance, pixels that straddle an occluding boundary correspond to large depth gaps in the scene. While the lidar may not measure every pixel, if we know it projects onto a tree, trees tend to stand out from the ground, which informs the topology of the scene. On the other hand, pixels that straddle illumination boundaries, like shadows cast by trees, seldom correspond to large depth discontinuities.

Depth completion is the process of assigning a depth value to each pixel. While there are several deep learning-based methods to do so, we wish to have the entire posterior estimate over depths. Sparse range measurements serve to ground the posterior estimate in a metric space. This could then be used by a decision and control engine downstream.
 
\begin{figure}[ht]
  \centering
  \includegraphics[width=0.5\textwidth]{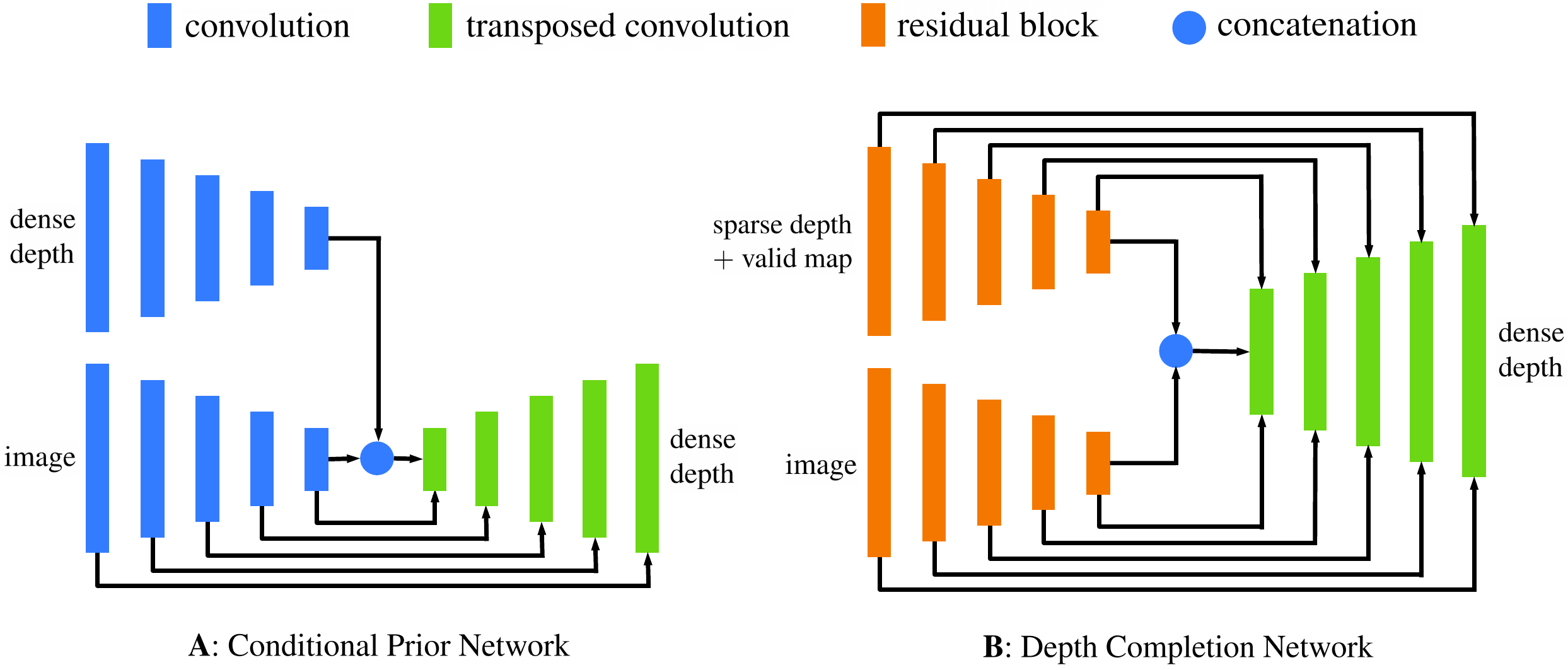}
  \vspace{-1.2em}
  \setlength{\belowcaptionskip}{-15pt}
  \caption{(A): the architecture of the Conditional Prior Network (CPN) to learn the conditional of the dense depth given a single image. (B): Our proposed Depth Completion Network (DCN) for learning the mapping from a sparse depth map and an image to a dense depth map. Connections within each encoder/decoder block are omitted for simplicity.}
  \label{fig:CPNDC}
\end{figure}

{\bf Side information.} If the dense depth map is obtained by processing the given image and sparse point cloud alone, the quality of the resulting decision or control action could be no better than if the raw data was fed downstream (Data Processing Inequality). However, if depth completion can exploit a prior or aggregate experience from previously seen images and corresponding dense depth maps, then it is possible for the resulting dense depth map to improve the quality of the decision or action, assuming that the training set is representative. To analyze a depth completion algorithm, it is important to understand what prior assumptions, hypotheses or side information is being exploited.

{\bf Goal.} We seek methods to {\em estimate the geometry and topology of the scene given an image, a sparse depth map, and a body of training data consisting of images and the associated dense depth maps.} Our assumption is that the distribution of seen images and corresponding depth maps is representative of the present data (image and sparse point cloud) once restricted to a sparse domain.

Our method yields the full posterior over depth maps, which is much more powerful than any point estimate. For instance, it allows reasoning about confidence intervals. We elect the simplest point estimate possible, which is the maximum, to evaluate the accuracy of the posterior. It should be noted, however, that when there are multiple hypotheses with similar posterior, the point estimate could jump from one mode to another, and yet the posterior being an accurate representation of the unknown variable. More sophisticated point estimators, for instance, taking into account memory, or spatial distribution, non-maximum suppression, etc. could be considered, but here we limit ourselves to the simplest one.

{\bf Key idea.} While an image alone is insufficient to determine a depth map, certain depth maps are more probable than others given the image and a previously seen dataset. The key to our approach is a conditional prior model $P(d | I, {\cal D})$ that scores the compatibility of each dense depth map $d$ with the given image $I$ based on the previously observed dataset ${\cal D}$. This is computed using a Conditional Prior Network (CPN)  \cite{yang2018conditional} in conjunction with a model of the likelihood of the observed sparse point cloud $z$ under the hypothesized depth map $d$, to yield the posterior probability and, from it, a maximum a-posteriori (MAP) estimate of the depth map for benchmark evaluation: 
\begin{equation}
    \hat d = \arg\max_d P(d | I, z) \propto P(z | d ) P_{\cal D}(d | I).
\end{equation}
Let $D \subset {\mathbb R}^2$ be the image domain, sampled on a regular lattice of dimension $N\times M$, $I:D \rightarrow {\mathbb R}^3$ is a color image, with the range quantized to a finite set of colors, $d: D \rightarrow {\mathbb R}+$ is the dense depth map defined on the lattice $D$, which we represent with an abuse of notation as a vector of dimension $MN$: $d \in {\mathbb R}_+^{NM}$. $\Omega \subset D$ is a sparse subset of the image domain, with cardinality $K = |\Omega|$, where the function $d$ takes values $d(\Omega) = z \in {\mathbb R}_+^K$. Finally, ${\cal D} = \{d_j, I_j \}_{j = 1}^n$ is a dataset of images $I_j$ and their corresponding dense depth maps $d_j \in {\mathbb R}_+^{NM}$. Since we do not treat ${\cal D}$ as a random variable but a given set of data, we write it as a subscript. In some cases, we may have additional data available during training, for instance stereo imagery, in which case we include it in the dataset, and discuss in detail how to exploit it in Sect.~\ref{sect:stereo}.

{\bf Results.} We train a deep neural network model to produce an estimate of the posterior distribution of dense depth maps given an image and a sparse point cloud (sparse range map), that leverages a Conditional Prior Network to restrict the hypothesis space, weighted by a classical likelihood term. We use a simple maximum a-posteriori (MAP) estimate to evaluate our approach on benchmark datasets, including the KITTI-unsupervised, where the dense depth map is predicted given an image and a point cloud with 5\% pixel coverage, and the KITTI-supervised, where a point cloud with 30\% coverage is given for training. We achieve top performance in both. We also validate on additional data in the Supplementary Materials \cite{yang2019dense}.

\vspace{-0.4em}
\section{Related Work}
\vspace{-0.4em}
\textbf{Semi-Dense Depth Completion.}
Structured light sensors typically provide dense depth measurements with about $20\%$ missing values; At this density, the problem is akin to inpainting \cite{camplani2012efficient,lu2014depth,shen2013layer} that use morphological operations  \cite{Ku2018InDO,premebida2016high}. The regime we are interested in involves far sparser point clouds ($>90\%$ missing values).

\textbf{Supervised Depth Completion.}
Given a single RGB image and its associated sparse depth measurements along with dense ground truth, learning-based methods \cite{eldesokey2018propagating,huang2018hms,ren2018sbnet,Uhrig2017SparsityIC,zhang2018deep} minimize the corresponding loss between prediction and ground truth depth. \cite{Uhrig2017SparsityIC} trains a deep network to regress depth using a sparse convolutional layer that discounts the invalid depth measurements in the input while \cite{huang2018hms} proposes a sparsity-invariant upsampling layer, sparsity-invariant summation, and joint sparsity-invariant concatenation and convolution. \cite{eldesokey2018propagating} treats the binary validity map as a confidence map and adapts normalized convolution for confidence propagation through layers. \cite{dimitrievski2018learning} implements an approximation of morphological operators using the contra-harmonic mean (CHM) filter \cite{masci2013learning} and incorporates it as a layer in a U-Net architecture for depth completion. \cite{chodosh2018deep} proposes a deep recurrent auto-encoder to mimic the optimization procedure of compressive sensing for depth completion, where the dictionary is embedded in the neural network. \cite{zhang2018deep} predicts surface normals and occlusion boundaries from the RGB image, which gives a coarse representation of the scene structure. The predicted surface normals and occlusion boundaries are incorporated as constraints in a global optimization framework guided by sparse depth.

\textbf{Unsupervised Depth Completion.} In this problem setting, dense ground truth depth is not available as supervision, so a strong prior is key. \cite{ma2018self} proposes minimizing the photometric consistency loss among a sequence of images with a second-order smoothness prior based on a similar formulation in single image depth prediction \cite{mahjourian2018unsupervised,wang2018learning,zhou2017unsupervised}. Instead of having a separate pose network or using direct visual odometry methods, \cite{ma2018self} uses Perspective-n-Point (PnP) \cite{lepetit2009epnp} and Random Sample Consensus (RANSAC) \cite{fischler1981random} to obtain pose. We exploit recently introduced method to learn the conditional prior \cite{yang2018conditional} to take into account scene semantics rather than using a local smoothness assumption.

\textbf{Stereo as Supervision.} Recent works in view synthesis \cite{flynn2016deepstereo, xie2016deep3d} and unsupervised single image depth prediction, \cite{garg2016unsupervised, godard2017unsupervised} propose using view synthesis to hallucinate a novel view image by reconstruction loss. In the case of stereo pairs, \cite{garg2016unsupervised, godard2017unsupervised} propose training networks to predict the disparities of an input image by reconstructing the unseen right view of a stereo pair given the left. In addition to the photometric reconstruction loss, local smoothness is assumed; \cite{godard2017unsupervised} additionally proposed edge-aware smoothness and left-right consistency. Although during inference, we assume only one image is given, at training time we may have stereo imagery available, which we exploit as in Sect.~\ref{sect:stereo}. In this work, we incorporate only the stereo photometric reconstruction term. Despite our network predicting depths and the network \cite{garg2016unsupervised, godard2017unsupervised} predicting disparities, we are able to incorporate this training scheme seamlessly into our approach.

\textbf{Exploiting Semantics and Contextual Cues.}
While methods \cite{eldesokey2018propagating, huang2018hms, ma2018self, ren2018sbnet, Uhrig2017SparsityIC, zhang2018deep} learn a representation for the depth completion task through ground truth supervision, they do not have any explicit modeling of the semantics of the scene. Recently, \cite{schneider2016semantically} explored this direction by predicting object boundary and semantic labels through a deep network and using them to construct locally planar elements that serve as input to a global energy minimization for depth completion. \cite{cheng2018depth} proposes to complete the depth by anisotropic diffusion with a recurrent convolution network, where the affinity matrix is computed locally from an image. \cite{jaritz2018sparse} also trains a U-Net for joint depth completion and semantic segmentation in the form of multitask learning in an effort to incorporate semantics in the learning process.

To address contextual cues and scene semantics, \cite{yang2018conditional} introduces a Conditional Prior Network (CPN) in the context of optical flow, which serves as a learning scheme for inferring the distribution of optical flow vectors given a single image. We leverage this technique and formulate depth completion as a maximum a-posteriori problem by factorizing it into a likelihood term and a conditional prior term, making it possible to explicitly model the semantics induced regularity of a single image. Even though our method could be applied to sparse-to-dense interpolation for optical flow, where the sparse matches can be obtained using \cite{yang2017s2f,yang2015coarse}, here we focus our test on depth completion task.

\vspace{-0.4em}
\section{Method}
\vspace{-0.4em}
\label{sect:methodology}

In order to exploit a previously observed dataset ${\cal D}$, we use a Conditional Prior Network (CPN) \cite{yang2018conditional} in our framework. Conditional Prior Networks infer the probability of an optical flow given a single image. During training, ground truth optical flow is encoded (upper branch in Fig.~\ref{fig:CPNDC}-A), concatenated with the encoder of an image (lower branch), and then decoded into a reconstruction of the input optical flow.

In our implementation, the upper branch encodes dense depth, concatenated with the encoding of the image, to produce a dense reconstruction of depth at the decoder, together with a normalized likelihood that can serve as a posterior score. We consider a CPN as a function that, given an image (lower branch input) maps any sample putative depth map (upper branch input) to a positive real number, which represents the conditional probability/prior of the input dense depth map given the image.

We denote the ensemble of parameters in the CPN as $w^{CPN}$; with abuse of notation, we denote the decoded depth with $d'=w^{CPN}(d, I)$. When trained  with a bottleneck imposed on the encoder (upper branch), the reconstruction error is  proportional to the conditional distribution:
\begin{equation}
    Q(d, I; w^{CPN}) = e^{-\| w^{CPN}(d,I) - d \|^{\eta}} \propto P_{{\cal D}}(d|I)
    \label{eq:CPN}
\end{equation}
where $\eta$ indicates the specific norm used for calculating $Q$. In Sect.~\ref{sect:training-procedure} and Sect.~\ref{experiment}, we show the training details of CPN, and also quantitatively show the effect of different choices of the norm $\eta$. In the following, we assume $w^{CPN}$ is trained, and $Q$ will be used as the conditional prior. For the proof that $Q$ computed by CPN represents the conditional prior as in Eq.~\eqref{eq:CPN}, please refer to \cite{yang2018conditional}.

In order to obtain a posterior estimate of depth, the CPN needs to be coupled with a likelihood term.

\subsection{Supervised Single Image Depth Completion}
\label{sect:supervised-depth-completion}
Supervised learning of dense depth assumes the availability of ground truth dense depth maps. In the KITTI depth completion benchmark \cite{Uhrig2017SparsityIC}, these are generated by accumulating the neighboring sparse lidar measurements. Even though it is called ground truth, the density is only $\sim 30\%$ of the image domain, whereas the density of the unsupervised benchmark is $\sim 5\%$. The training loss in the supervised modality is just the prediction error:
\begin{equation}
    L(w) = \sum_{j=1}^{N} \| \phi(z_j,I_j;w) - d_j \|^{\gamma}
    \label{eq:sup-loss}
\end{equation}
where $\phi$ is the map from sparse depth $z$ and image $I$ to dense depth, realized by a deep neural network with parameters $w$, and $\gamma = 1$ fixed in the supervised training.

Our network structure for $\phi$ is detailed in Fig. \ref{fig:CPNDC}-B, which has a symmetric two-branch structure, each encoding different types of input: one sparse depth, the other an image; skip connections are enabled for two branches. Note that our network structure is unique among all the top performing ones on the KITTI depth completion benchmark: We do not use specifically-designed layers for sparse inputs, such as sparsity invariant layers \cite{huang2018hms,Uhrig2017SparsityIC}. Instead of early fusion of sparse depth and image, our depth defers fusion to decoding, which entails fewer learnable parameters, detailed in \cite{yang2019dense}. A related idea is proposed in \cite{jaritz2018sparse}; instead of a more sophisticated NASNet block \cite{zoph2017learning}, we use the more common ResNet block \cite{he2016deep}. Although simpler than competing methods, our network achieves state-of-the-art performance (Sect.~\ref{experiment}).

\vspace{-0.4em}
\subsection{Unsupervised Single Image Depth Completion}
\vspace{-0.4em}
Supervised learning requires ground truth dense depth, which is hard to come by. Even the ``ground truth'' provided in the KITTI benchmark is only $30\%$ dense and interpolated from even sparser maps. When only sparse independent measurements of depth are available, for instance from lidar, with less than $10\%$ coverage (e.g. $5\%$ for KITTI), we call depth completion {\em unsupervised} as the only input are sensory data, from images and a range measurement device, with no annotation or pre-processing of the data.

The key to our approach is the use of a CPN to score the compatibility of each dense depth map $d$ with the given image $I$ based on the previously observed data ${\cal D}$. In some cases, we may have additional sensory data available {\em during training}, for instance, a second image taken with a camera with a known relative pose, such as stereo. In this case, we include the reading from the second camera in the training set ${\cal D}$, as described in Sect.~\ref{sect:stereo}. When only a single image is given, the CPN Eq.~\eqref{eq:CPN} is combined with a model of the likelihood of the observed sparse point cloud $z$ under the hypothesized depth map $d$:
\begin{equation}
    P(z | d ) \propto e^{- \| z - d(\Omega) \|^{\gamma}}
    \label{eq:likelihood_sparse_depth}
\end{equation}
which is simply a Gaussian around the hypothesized depth, restricted to the sparse subset $\Omega$, when $\gamma=2$. The overall loss is:
\begin{multline}
    L^{u}(w) = - \sum_{j=1}^N log P(d_j | I_j, z_j, {\cal D}) \\
    = \sum_{j=1}^N \| z_j - d_j(\Omega) \|^{\gamma} + \alpha \sum_{j=1}^N \| w^{CPN}(d_j,I_j) - d_j \|^{\eta} \\
    = \sum_{j=1}^N \| z_j - \phi(z_j,I_j;w)(\Omega) \|^{\gamma} + \\ \alpha \sum_{j=1}^N \| w^{CPN}(\phi(z_j,I_j;w),I_j) - \phi(z_j,I_j;w) \|^{\eta}
    \label{eq:unsup-single-image-dc}
\end{multline}
Note that $\gamma, \eta$ control the actual norm used during training, as well as the modeling of the likelihood and conditional distribution. We experiment with these parameters in Sect.~\ref{subsec:norm}, and show our quantitative analysis there.

\subsection{Disparity Supervision} \label{sect:stereo}
Some datasets come with stereo imagery. We want to be able to exploit it, but without having to require its availability at inference time. We exploit the strong relation between depth and disparity. In addition to the sparse depth $z$ and the image $I$, we are given a second image $I'$ as part of a stereo pair, which is rectified (standard pre-processing), to first-order we assume that there exists a displacement $s = s(x), x \in D$ such that
\begin{equation}
    I(x) \approx I'(x+s)
    \label{eq:stereo-reconstruction}
\end{equation}
which is the intensity constancy constraint. We model, again simplistically, disparity $s$ as $s = FB/d$, where $F$ is the focal length and $B$ is the baseline (distance between the optical centers) of the cameras. Hence, we can synthesize disparity $s$ from the predicted dense depth $d$, thus to constrain the recovery of 3-d scene geometry. More specifically, we model the likelihood of seeing $I'$ given $I, d$ as:
\begin{equation}
    P(I'| I, d) \propto e^{-\dfrac{\sum_x \|I(x)-I'(x+s(d(x)))\| }{\delta^2}}
    \label{eq:photometric-loss-stereo}
\end{equation}

\begin{table}[t]
\centering
\footnotesize
\begin{tabular}{lllllll}
\toprule
Method & iRMSE & iMAE & RMSE & MAE & Rank \\ \midrule
Dimitrievski \cite{8569539} &
3.84 & 1.57 & 1045.45 & 310.49 & 13.0\\ \midrule
Cheng \cite{cheng2018depth} &
2.93 & 1.15 & 1019.64 & 279.46 & 7.5\\ \midrule
Huang \cite{huang2018hms} &
2.73 & 1.13 & 841.78 & 253.47 & 6.0\\ \midrule
Ma \cite{ma2018self}  & 
2.80 & 1.21 & \textbf{814.73} & 249.95 & 5.5\\ \midrule
Eldesokey \cite{eldesokey2018propagating} & 
2.60 & 1.03 & 829.98 & 233.26 & 4.75\\ \midrule
Jaritz \cite{jaritz2018sparse} &
2.17 & 0.95 & 917.64 & 234.81  & 3.0\\ \midrule
Ours & \textbf{2.12} & \textbf{0.86} & 836.00 & \textbf{205.40} & \textbf{1.5}
\\ \bottomrule
\end{tabular}
\setlength{\belowcaptionskip}{-10pt}
\caption{Quantitative results on the supervised KITTI depth completion benchmark. Our method achieves state of the art performance in three metrics, iRMSE, iMAE, and MAE. \cite{ma2018self} performs better than us by 2.6\% on the RMSE metric; however, we outperform \cite{ma2018self} on all other metrics by 24.3\%, 28.9\% and 17.8\% on the iRMSE, iMAE and MAE, respectively. The last column is the average rank over ranks on all the four metrics.}
\label{tab:supervised-results}
\end{table}

However, the validity of the intensity constancy assumption is affected by complex phenomena such as translucency, transparency, inter-reflection, etc. In order to mitigate the error in the assumption, we could also employ a perceptual metric of structural similarity (SSIM) \cite{wang2004image}. SSIM scores corresponding $3 \times 3$ patches $p(x), p'(x) \in \mathbb{R}_+^{3 \times 3}$ centered at $x$ in $I$ and $I'$, respectively, to measure their local structural similarity. Higher scores denote more similarity; hence we can subtract the scores from 1 to form a robust version of Eq.~\eqref{eq:photometric-loss-stereo}. We use $P_{raw}(I'|I,d)$ and $P_{ssim}(I'|I,d)$ to represent the probability of $I'$ given $I, d$ measured in raw photometric value and SSIM score respectively. When the stereo pair is available, we can form the conditional prior as follows by applying conditional independence:
\begin{multline}
    P(d|I, I', {\cal D}) \propto P(I'|I, d, {\cal D})P(d|I, {\cal D}) \\
    = P(I'|I, d)P_{\cal D}(d|I)
\end{multline}
Similar to the training loss Eq.~\eqref{eq:unsup-single-image-dc} for the unsupervised single image depth completion setting, we can derive the loss for the stereo setting as follows:
\begin{multline}
    L^{s}(w) = - \sum_{j=1}^N log P(d_j | I_j, I'_j, z_j, {\cal D}) \\
    = L^{u}(w) + \beta \sum_{j,x} \|I_j(x)-I'_j(x+s(d_j(x)))\|
    \label{eq:unsup-stereo-image-dc}
\end{multline}
where $d_j = \phi(z_j,I_j;w)$ and $L^{u}$ is the loss defined in Eq.~\eqref{eq:unsup-single-image-dc}. Note that, the above summation term is the instantiation for $P_{raw}(I'|I,d)$, which can also be replaced by the SSIM counterpart. Rather than choosing one or the other, we compose the two with tunable parameters $\beta_c$ and $\beta_s$, our final loss for stereo setting depth completion is:
\begin{equation}
    \label{eq:final-loss-for-stereo}
    L^{s}(w) = L^{u}(w) + \beta_c\psi_c + \beta_s\psi_s
\end{equation}
with $\psi_c$ represents the raw intensity summation term in Eq.~\eqref{eq:unsup-stereo-image-dc}, and $\psi_s$ for the SSIM counterpart. Next, we elaborate our implementation details and evaluate the performance of our proposed method in different depth completion settings.

\vspace{-0.4em}
\section{Implementation Details}
\vspace{-0.4em}
\label{sect:implementation}

\subsection{Network architecture}

We modify the public implementation of CPN \cite{yang2018conditional} by replacing the input of the encoding branch with a dense depth map. Fusion of the two branches is simply a concatenation of the encodings. The encoders have only convolutional layers, while the decoder is made of transposed convolutional layers for upsampling.

Our proposed network, unlike the base CPN, as seen in Fig.~\ref{fig:CPNDC}-A, contains skip connections between the layers of the depth encoder and the corresponding decoder layers, which makes the network symmetric. We also use ResNet blocks \cite{he2016deep} in the encoders instead of pure convolutions. A stride of $2$ is used for downsampling in the encoder and the number of channels in the feature map after each encoding layer is $[64*k, 128*k, 256*k, 512*k, 512*k]$. In all our experiments, we use $k=0.25$ for the depth branch, and $k=0.75$ for the image branch, taking into consideration that an RGB image has three channels while depth map only has one channel. Our network has fewer parameters than those based on early fusion (e.g.  \cite{ma2018self} used $\approx$27.8M parameters in total; where as we only use $\approx$18.8M). We provide an example comparing our network architecture and that of \cite{ma2018self} in the Supplementary Materials \cite{yang2019dense}.

\begin{table}[t]
\begin{adjustwidth}{-.25in}{-.00in}
\footnotesize
\begin{tabular}{lllllll}
\toprule
 & \multicolumn{2}{c}{Validation Set} & \multicolumn{4}{c}{Test Set} \\
\midrule 
Loss & RMSE & MAE & iRMSE & iMAE & RMSE & MAE \\ \midrule
Ma \cite{ma2018self} & 1384.85 & 358.92 & 4.07 & 1.57 & 1299.85 & 350.32 \\ \midrule
$L^u$ & 1325.79 & 355.86 & 3.69 & 1.37 & 1285.14 & 353.16  \\ \midrule
$L^s(\psi_c)$ & 1320.26 & 353.24 &  3.63 & 1.34 & 1274.65 & 349.88 \\ \midrule
$L^s(\psi_c,\psi_s)$ & \textbf{1310.03} & \textbf{347.17} & \textbf{3.58} & \textbf{1.32} & \textbf{1263.19} & \textbf{343.46} \\ \bottomrule
\end{tabular}
\end{adjustwidth}
\setlength{\belowcaptionskip}{-25pt}
\caption{Quantitative results on the unsupervised KITTI depth completion benchmark. Our baseline approach using CPN as a regularizer outperforms \cite{ma2018self} on the iRMSE, iMAE and RMSE metrics on the test set, whereas \cite{ma2018self} marginally performs better than us on MAE by 0.8\%. We note that \cite{ma2018self} achieves this performance using photometric supervision. When including our photometric term (Eq.~\eqref{eq:final-loss-for-stereo}), we outperform \cite{ma2018self} on every metric and achieve state-of-the-art performance.}
\vspace{-1.2em}
\label{tab:unsupervised-results}
\end{table}

\vspace{-0.4em}
\subsection{Training Procedure}
\vspace{-0.4em}
\label{sect:training-procedure}
We begin by detailing the training procedure for CPN. Once learned, we apply CPN as part of our training loss and do not need it during inference. In order to learn the conditional prior of the dense depth maps given an image, we require a dataset with images and corresponding dense depth maps. We are unaware of any real-world dataset for outdoor scenes that meets our criterion. Therefore, we train the CPN using the Virtual KITTI dataset \cite{Gaidon:Virtual:CVPR2016}. It contains 50 high-resolution monocular videos with a total of $21,260$ frames, together with ground truth dense depth maps, generated from five different virtual worlds under different lighting and weather conditions. The original Virtual KITTI image has a large resolution of $1242\times 375$, which is too large to feed into a normal commercial GPU. So we crop it to $768\times 320$ and use a batch size of 4 for training. The initial learning rate is set to $1e^{-4}$, and is halved every 50,000 steps 300,000 steps in total.

\begin{figure}[]
  \centering
  \includegraphics[width=0.35\textwidth]{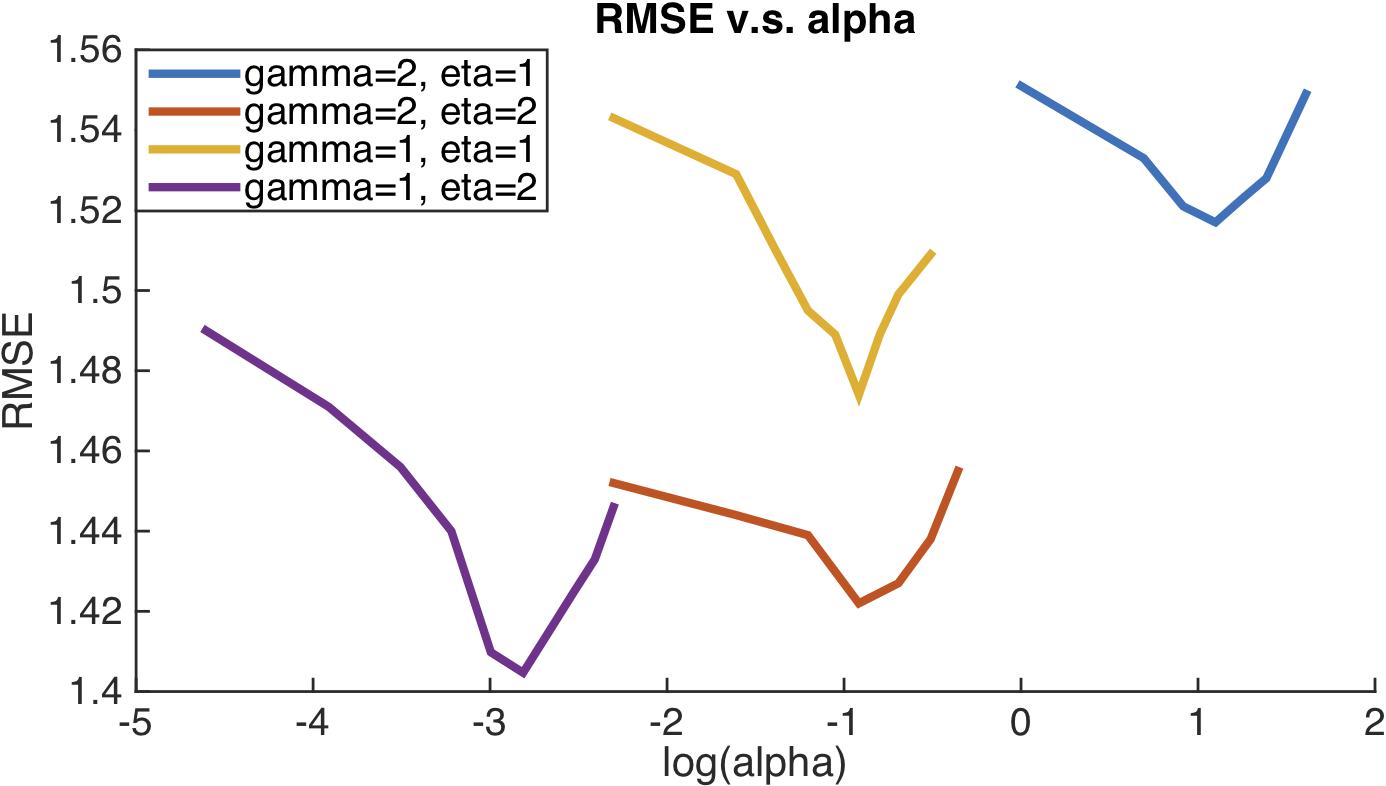}
  \setlength{\belowcaptionskip}{-12pt}
  \caption{This plot shows the empirical study on the choice of norms $\gamma, \eta$ in the likelihood term and the conditional prior term respectively. Each curve is generated by varying $\alpha$ in Eq.~\eqref{eq:unsup-single-image-dc} with fixed $\gamma, \eta$. And the performance is measured in RMSE.}
  \label{fig:rmse-alpha}
\end{figure}

We implement our approach using TensorFlow \cite{abadi2016tensorflow}. We use Adam \cite{kingma2014adam} to optimize our network with the same batch size and learning rate schedule as the training of CPN. We apply histogram equalization and also randomly crop the image to $768\times 320$. We additionally apply random flipping both vertical and horizontal to prevent overfitting. In the case of unsupervised training, we also perform a random shift within a $3 \times 3$ neighborhood to the sparse depth input and the corresponding validity map. We use $\alpha=0.045$, $\beta=1.20$ for Eq.~\eqref{eq:unsup-stereo-image-dc}, and the same $\alpha$ is applied with $\beta_c=0.15$, $\beta_s=0.425$ for Eq.~\eqref{eq:final-loss-for-stereo}. We choose $\gamma=1$ and $\eta=2$, but as one may notice in Eq.~\eqref{eq:CPN}, the actual conditional prior also depends on the choice of the norm $\eta$. To show the reasoning behind our choice, we will present as an empirical study in Fig.~\ref{fig:rmse-alpha} to show the effects of the different pairing of norms with a varying $\alpha$ by evaluating each model on the RMSE metric.

In the next section, we report representative experiments in both the supervised and unsupervised benchmarks.

\vspace{-0.2em}
\section{Experiments}
\label{experiment}
We evaluate our approach on the KITTI depth completion benchmark \cite{Uhrig2017SparsityIC}. The dataset provides $\sim 80k$ raw image frames and corresponding sparse depth maps. The sparse depth maps are the raw output from the Velodyne lidar sensor, each with a density of about $5\%$. The ground truth depth map is created by accumulating the neighboring 11 raw lidar scans, with roughly $30\%$ pixels annotated. We use the officially selected 1,000 samples for validation and we apply our method to 1,000 testing samples, with which we submit to the official KITTI website for evaluation. We additionally perform an ablation study on the effects of the sparsity of the input depth measurements on the NYUv2 indoor dataset \cite{silberman2012indoor} in the Supplementary Materials \cite{yang2019dense}.

\vspace{-0.4em}
\subsection{Norm Selection}
\label{subsec:norm}
As seen in Eq.~\eqref{eq:unsup-single-image-dc}, $\gamma, \eta$ control the actual norms (penalty functions) applied to the likelihood term and conditional prior term respectively, which in turn determine how we model the distributions. General options are from the binary set $\{ 1, 2 \}$. i.e. $\{ {\cal L}_1, {\cal L}_2 \}$, however, there is currently no agreement on which one is better suited for the depth completion task. \cite{ma2018self} shows $\gamma = 2$ gives significant improvement for their network, while both \cite{Uhrig2017SparsityIC,jaritz2018sparse} claim to have better performance when $\gamma = 1$ is applied. In our approximation of the posterior in Eq.~\eqref{eq:unsup-single-image-dc}, the choice of the norms gets more complex as the modeling (norm) of the conditional prior will also depend on the likelihood model. Currently, there is no clear guidance on how to make the best choice, as it may also depend on the network structure. Here we try to explore the characteristic of different norms, at least for our network structure, by conducting an empirical study on a simple version (channel number of features reduced) of our depth completion network using different combinations of $\gamma$ and $\eta$. As shown in Fig.~\ref{fig:rmse-alpha}, the performance on the KITTI depth completion validation set varies in a wide range with different $\gamma, \eta$. Clearly for our depth completion network, ${\cal L}_1$ is always better than ${\cal L}_2$ on the likelihood term. And the lowest RMSE is achieved when a ${\cal L}_2$ is also applied on the conditional prior term. Thus the best coupling is $\gamma=1, \eta=2$ for Eq.~\eqref{eq:unsup-single-image-dc}.

\begin{figure*}[!ht]
  \centering
  \includegraphics[width=0.91\textwidth]{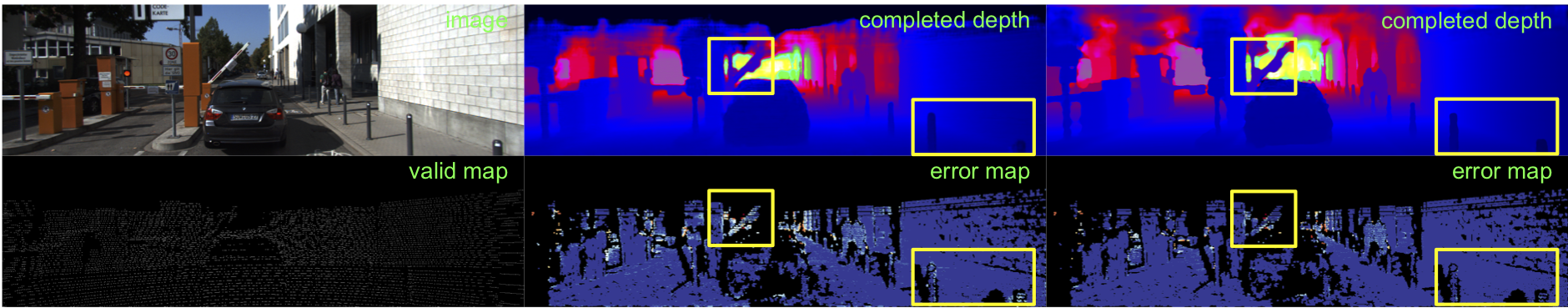}
  \includegraphics[width=0.91\textwidth]{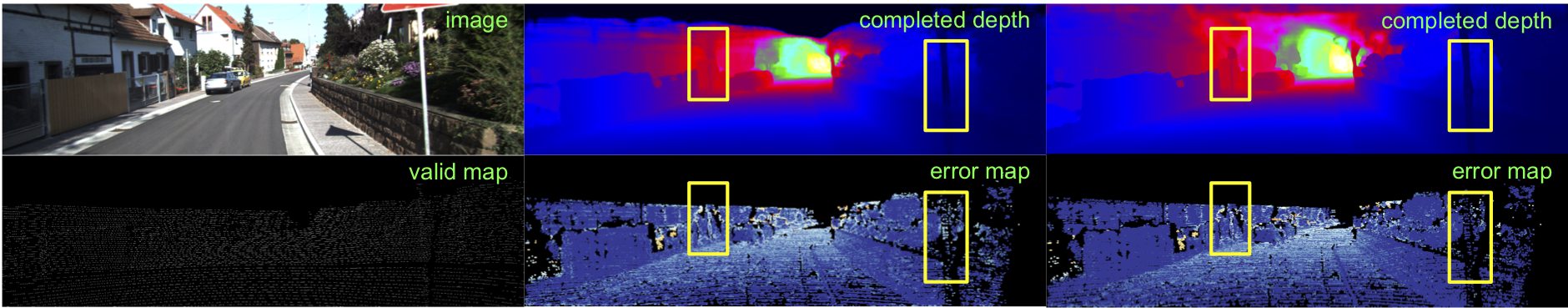}
  \includegraphics[width=0.91\textwidth]{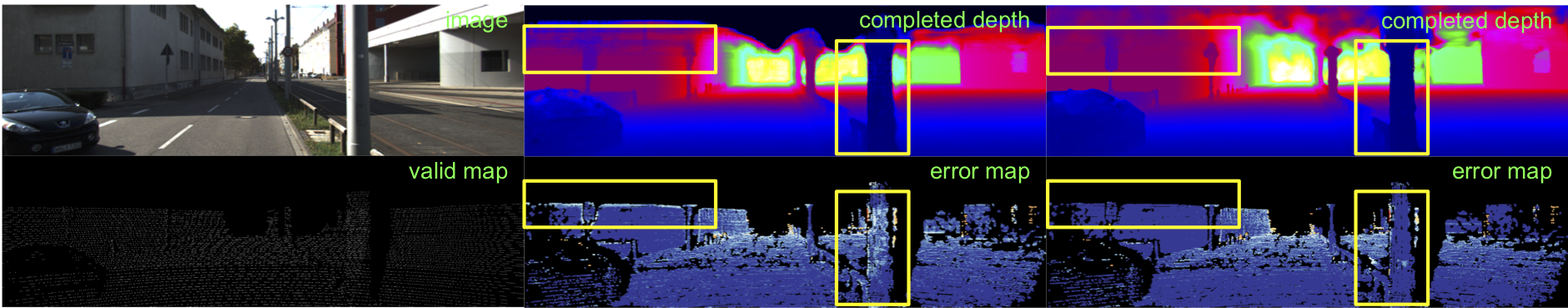}
  \includegraphics[width=0.91\textwidth]{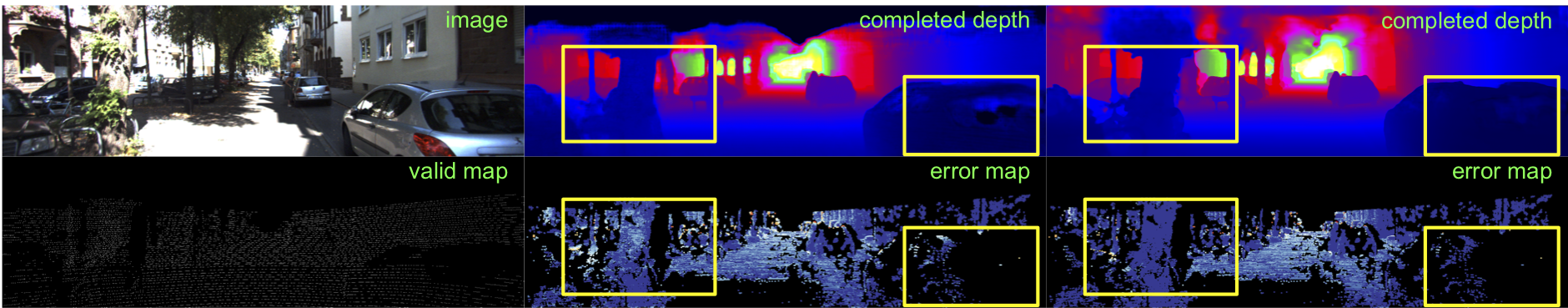}
  \caption{Qualitative comparison to Ma et al. \cite{ma2018self} on KITTI depth completion test set in the supervised setting. Image and validity map of the sparse measurements (1st column), dense depth results and corresponding error map of \cite{ma2018self} (2nd column) and our results and error map (3rd column). Warmer color in the error map denotes higher error. The yellow rectangles highlight the regions for detailed comparison. Note that our network consistently performs better on fine and far structures and our completed dense depth maps have less visual artifacts.}
  \label{fig:visual-kitti-sup}
\end{figure*}

\vspace{-0.2em}
\subsection{Supervised Depth Completion}
We evaluate the proposed Depth Completion Network described in Sect.~\ref{sect:supervised-depth-completion} on the KITTI depth completion benchmark. We show a quantitative comparison between our approach and the top performers on the benchmark in Tab.~\ref{tab:supervised-results}. Our approach achieves the state-of-the-art in three metrics by outperforming  \cite{eldesokey2018propagating,  jaritz2018sparse}, who each held the state-of-the-art in different metrics on the benchmark. We improve over \cite{jaritz2018sparse} in iRMSE and iMAE by 2.3\% and 9.5\%, respectively, and \cite{eldesokey2018propagating} in MAE by 11.9\%. \cite{ma2018self} performs better on the RMSE metric by 2.6\%; however, we outperform \cite{ma2018self} by 24.3\%, 28.9\% and 17.8\% on the iRMSE, iMAE and MAE metrics, respectively. Note in the online table of KITTI depth completion benchmark\footnote{\url{http://www.cvlibs.net/datasets/kitti/eval_depth.php?benchmark=depth_completion}}, all methods are solely ranked by the RMSE metric, which may not fully reflect the performance of each method. Thus we propose to rank all methods by averaging over the rank numbers on each metric, and the overall ranking is shown in the last column of Tab.~\ref{tab:supervised-results}. Not surprisingly, our depth completion network gets the smallest rank number due to its generally good performance on all metrics.

Fig.~\ref{fig:visual-kitti-sup} shows a qualitative comparison of our method to the top performing method on the test set of the KITTI benchmark. We see that our method produces depths that are more consistent with the scene with fewer artifacts (e.g. grid-like structures \cite{ma2018self}, holes in objects \cite{eldesokey2018propagating}). Also, our network performs consistently better on fine and far structures, which may be traffic signs and poles on the roadside that provide critical information for safe driving as shown in the second row in Fig.~\ref{fig:visual-kitti-sup}. More in the Supplementary \cite{yang2019dense}.

\begin{figure*}[!ht]
  \centering
  \includegraphics[width=0.91\textwidth]{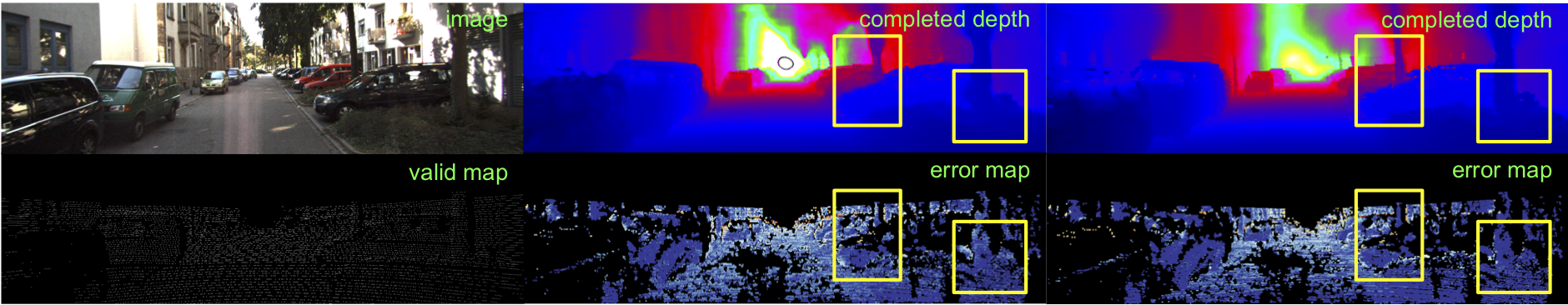}
  \includegraphics[width=0.91\textwidth]{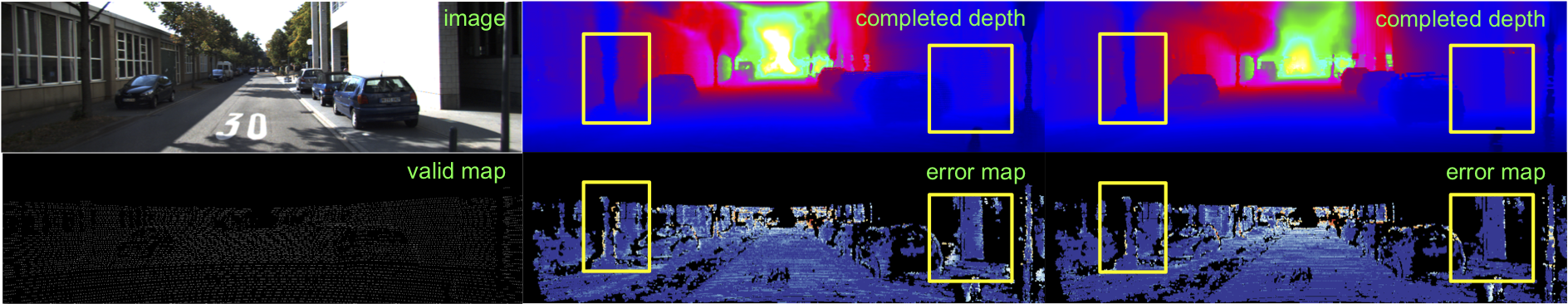}
  \includegraphics[width=0.91\textwidth]{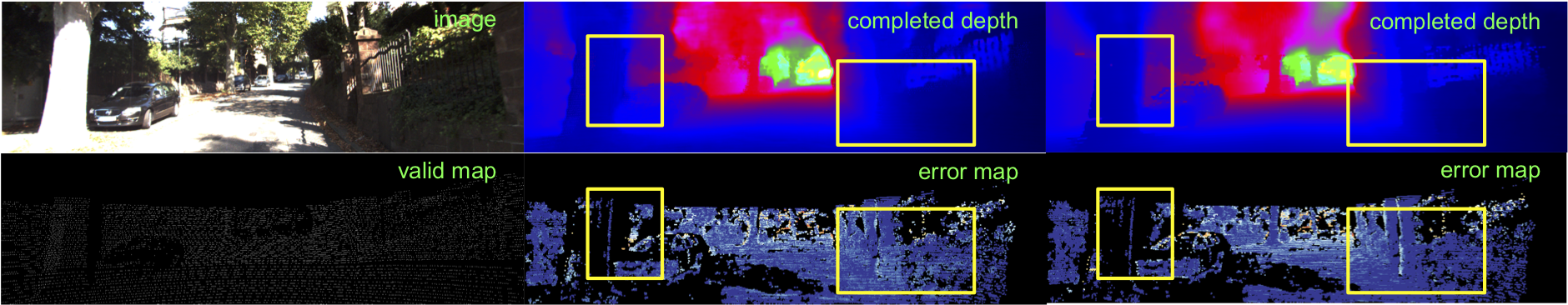}
  \caption{Qualitative comparison to Ma et al. \cite{ma2018self} on the KITTI depth completion test set in the unsupervised setting. Image and validity map of the sparse measurements (1st column), dense depth results and corresponding error map of \cite{ma2018self} (2nd column) and ours (3rd column). Warmer color in the error map denotes higher error. Yellow rectangles highlight the regions for detailed comparison. Note again that our network consistently performs better on fine and far structures and our completed dense depth maps have less visual artifacts (this includes the circle in the center of their prediction, row 1, column 2).}
  \label{fig:visual-kitti-unsup}
\end{figure*}

\vspace{-0.4em}
\subsection{Unsupervised Depth Completion}
We show that our network can also be applied to unsupervised setting using only the training loss Eq.~\eqref{eq:unsup-single-image-dc} to achieve the state-of-the-art results as well. We note that the simplest way for the network to minimize the data term is to directly copy the sparse input to the output, which will make the learning inefficient. To facilitate the training, we change the stride of the first layer from 1 to 2 and replace the final layer of the decoder with a nearest neighbor upsampling.

We show a quantitative comparison (Tab.~\ref{tab:unsupervised-results}) between our method and that of \cite{ma2018self} along with an ablation study on our loss function. We note that the results of \cite{ma2018self} are achieved using their full model, which includes their multi-view photometric term. Our approach using just Eq.~\eqref{eq:unsup-single-image-dc} is able to outperform \cite{ma2018self} in every metric with the exception of MAE where \cite{ma2018self} marginally beats us by 0.8\%. By applying our reconstruction loss Eq.~\eqref{eq:unsup-stereo-image-dc}, we outperform \cite{ma2018self} in every metric. Moreover, our full model Eq.~\eqref{eq:final-loss-for-stereo} further improves over all other variants and is state-of-the-art in unsupervised depth completion. We present a qualitative comparison between our approach and that of \cite{ma2018self} in Fig.~\ref{fig:visual-kitti-unsup}. Visually, we observe the results of \cite{ma2018self} still contain the artifacts as seen before. The artifacts, i.e. circles, as detailed in Fig.~\ref{fig:visual-kitti-unsup}, are signs that their network is probably overfitted to the input sparse depth, due to the lack of semantic regularity. Our approach, however, does not suffer from these artifacts; instead, our predictions are globally correct and consistent with the scene geometry.

\vspace{-0.4em}
\section{Discussion}
\vspace{-0.4em}
In this work, we have described a system to infer a posterior probability over the depth of points in the scene corresponding to each pixel, given an image and a sparse aligned point cloud. Our method leverages a Conditional Prior Network, that allows the association of a probability to each depth value based on a single image, and combines it with a likelihood term for sparse depth measurements. Moreover, we exploit the availability of stereo imagery in constructing a photometric reconstruction term that further constrains the predicted depth to adhere to the scene geometry.

We have tested the approach both in a supervised and unsupervised setting. It should be noted that the difference between ``supervised'' and ``unsupervised'' in the KITTI benchmark is more quantitative than qualitative: the former has about $30\%$ coverage in depth measurements, the latter about $5\%$. We show in Tab.~\ref{tab:supervised-results} and \ref{tab:unsupervised-results} that our method achieves state-of-the-art performance in both supervised and unsupervised depth completion on the KITTI benchmark. Although we outperform other methods on score metrics that measures the deviation from the ground truth, we want to emphasize that our method does not simply produce a point estimate of depth, but provides a confidence measure, that can be used for more downstream processing, for instance for planning, control and decision making.

We have explored the effect of various hyperparameters, and are in the process of expanding the testing to real-world environments, where there could be additional errors and uncertainty due to possible time-varying misalignment between the range sensor and the camera, or between the two cameras when stereo is available, faulty intrinsic camera calibration, and other nuisance variability inevitably present on the field that is carefully weeded out in evaluation benchmarks such as KITTI. This experimentation is a matter of years, and well beyond the scope of this paper. Here we have shown that a suitably modified Conditional Prior Network can successfully transfer knowledge from prior data, including synthetic ones, to provide context to input range values for inferring missing data. This is important for downstream processing as the context can, for instance, help differentiate whether gaps in the point cloud are free space or photometrically homogeneous obstacles, as discussed in our motivating example in Fig.~\ref{fig:task-illustration}.

\section{Supplementary}

\subsection{An Ablation Study on NYUv2 Indoor Dataset}

The goal of this ablation study is to see how the performance of our depth completion network changes while we vary its input depth density. By density we mean the number of valid depth measurements divided by the total number of pixels in an image. We choose 100k samples from the training set of NYUv2~\cite{silberman2012indoor} indoor dataset as our training set, and we separately select 500 samples as the test set. Each of the sample has dimension 480x640. During training we use a batch size of 1, and an initial learning rate of 1e-4, which is halved every 25000 steps until 100000 total number of steps. The performance with respect to the given density reported in Table~\ref{tab:nyuv2} is obtained at the end of the training, and the density is in percentage. As shown in Table~\ref{tab:nyuv2}, we also observe the performance degeneration phenomenon as observed in others \cite{huang2018hms,ma2018sparse}, which confirms again the negative effect of decreasing the density of the input sparse depth.

\begin{table}[h!]
  \scriptsize
  \centering
  \begin{tabular}{l c c c c c c c c c}
    \toprule
    density & 0.01 & 0.02 & 0.06 & 0.15 & 0.3 & 1.5 \\
    \midrule
    RMSE & 3.829 & 2.623 & 1.391 & 0.928 & 0.569 & 0.521 \\ \midrule
    AbsRel & 1.171 & 1.067 & 0.452 & 0.306 & 0.171 & 0.122 \\ \midrule
  \end{tabular}
  \caption{A quantitative evaluation of our proposed network structure on the NYUv2~\cite{silberman2012indoor} dataset with varying input sparse depth density. We denote density as percentage of input sparse depth with respect to the input image. The input sparse depth measurements were randomly sampled from the dense depth maps provided. It is apparent from the experiment that decreasing the density of the input depth causes the performance of the model to degrade.}
  \label{tab:nyuv2}
\end{table}

\subsection{Detail of $\psi_s$ in Eq.~\eqref{eq:final-loss-for-stereo}}

We state $P_{ssim}(I'|I,d)$ as the likelihood counterpart to Eq.~\eqref{eq:photometric-loss-stereo} for stereo setting. Here we show its detailed form. We denote $g(x,y; I, I')$ as the function to compute a SSIM score given two images $I, I'$ and two pixel locations $x, y$, each on different images. Then we have:
\begin{equation}
    P_{ssim}(I'|I,d) \propto e^{-\dfrac{\sum_{x} \| 1 - g(x, x+s(d(x)); I, I') \| }{\delta^2}}
    \label{eq:ssim-loss-stereo}
\end{equation}

with $d_j = \phi(z_j,I_j;w)$, the detailed loss as defined in Eq.~\eqref{eq:final-loss-for-stereo} is:
\begin{multline}
    \label{eq:final-loss-for-stereo-detailed}
    L^{s}(w) = L^{u}(w) + \beta_c\sum_{j,x} \|I_j(x)-I'_j(x+s(d_j(x)))\| \\
    + \beta_s\sum_{j,x} \| 1 - g(x, x+s(d_j(x)); I_j, I'_j) \|
\end{multline}

\subsection{On the Network Complexity}

In regards to the number of parameters used in the network versus the prior-art \cite{ma2018self}: Although our network has two branches that become fused at a later stage, we in fact have fewer parameters than that of \cite{ma2018self}. Here we use the second layer as an example. After early fusion, the input to the second layer for Ma et al. \cite{ma2018self} has dimension $b \times h \times w \times 64$, and the output of second layers in the network of Ma et al. has dimension $b \times \dfrac{h}{2} \times \dfrac{w}{2} \times 128$ so number of parameters used for the second layer is going to be $3 \times 3 \times 64 \times 128 = 73728$.

However, in our depth completion network, the sparse depth branch has input dimension $b \times h \times w \times 16$ and output dimension $b \times \dfrac{h}{2} \times \dfrac{w}{2} \times 32$, so the sparse depth branch has $3 \times 3 \times 16 \times 32 = 4608$ parameters. The image branch has input dimension $b \times h \times w \times 48$, output dimension $b \times \dfrac{h}{2} \times \dfrac{w}{2} \times 96$, so the image branch has $3 \times 3 \times 48 \times 96 = 41472$ parameters. In total, the second layer in our depth completion network contains only $46080$ trainable parameters. Much less that those in Ma et al \cite{ma2018self}. This same exercise can be applied similarly to other layers to compare the number of parameters used.

To summarize for the reader, \cite{ma2018self} contains a total of $\approx$27.8M parameters for their network; whereas, we only use $\approx$18.8M parameters.

\subsection{More Visual Comparisons}

\begin{figure*}[!ht]
  \centering
  \includegraphics[width=1.0\textwidth]{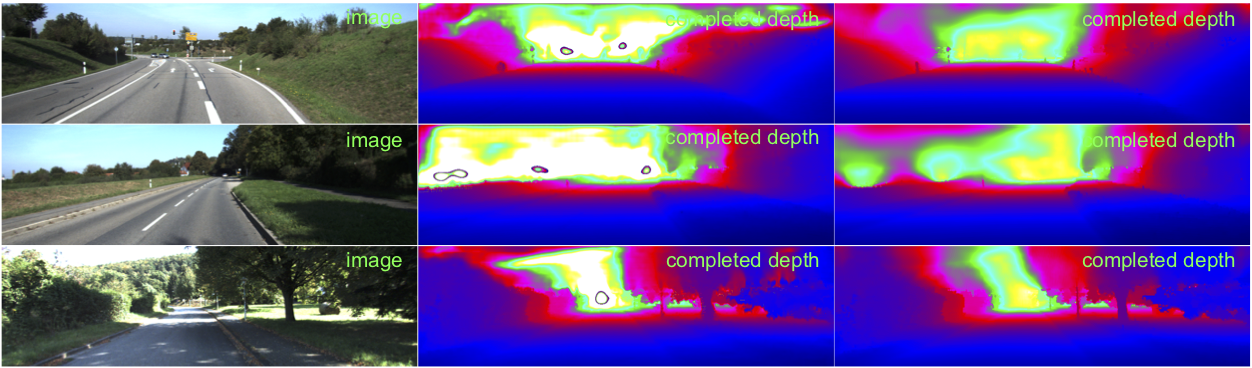}
  \setlength{\belowcaptionskip}{-25pt}
  \caption{Qualitative comparisons to Ma et al. \cite{ma2018self} in unsupervised setting. Note the ``O'' shape artifacts occur due to a lack of knowledge about the scene regularity in \cite{ma2018self}, which however is not present in our completed depth due to a more sophisticated scene prior.}
  \label{fig:artifacts-ma-unsup}
\end{figure*}

\begin{figure*}[!ht]
  \centering
  \includegraphics[width=1.0\textwidth]{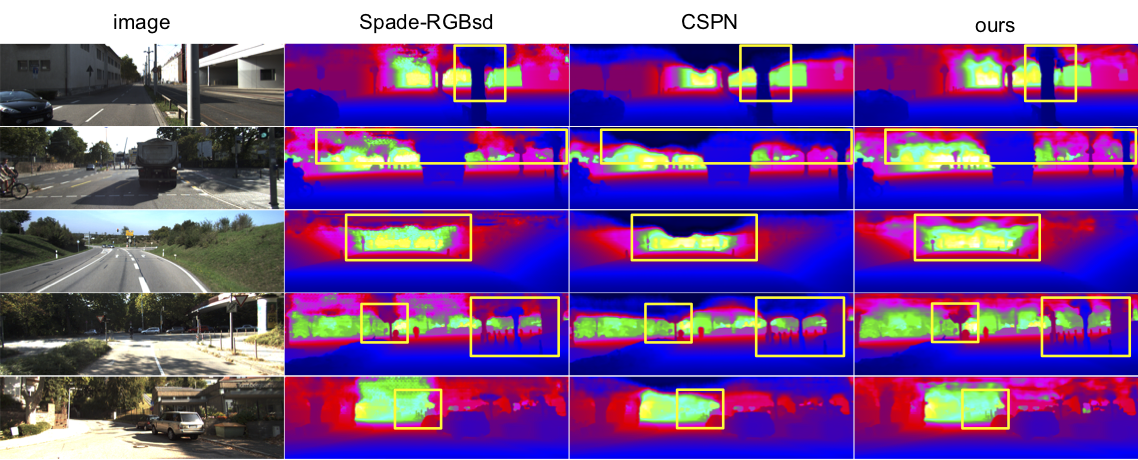}
  \setlength{\belowcaptionskip}{-25pt}
  \caption{More qualitative comparisons to Spade-RGBsD \cite{jaritz2018sparse} (second column) and CSPN \cite{cheng2018depth} (third column) in supervised setting. Yellow rectangles highlight detailed comparisons.}
  \label{fig:broad-sup}
\end{figure*}

In Fig.~\ref{fig:artifacts-ma-unsup}, we show more comparisons to the top method \cite{ma2018self} on KITTI depth completion benchmark in the unsupervised setting. Note the ``O'' shape artifacts in the results of Ma et al. \cite{ma2018self} were due to the lack of knowledge about the scene geometry. Such is not the case in our results as we employ a more sophisticated scene prior that discounts the unlikely artifacts that is not compatible with the scene depicted in the RGB image.

In Fig.~\ref{fig:broad-sup}, we show more comparisons to the state-of-the-art methods \cite{jaritz2018sparse, cheng2018depth} on the supervised depth completion task. Please see details therein.

\subsection{CPN Applied to Monocular Video}

Indeed our experiment is performed on the stereo setting; whereas video-based supervision, if properly leveraged, can exploit larger baselines. We have conducted the ablation study, shown in the table below. The use of CPN as a prior gives improvement (rows 1, 2), and our method in the monocular setting still outperforms \cite{ma2018self} (rows 3, 4).
\vspace{-0.6em}
\begin{table}[!h]
\centering
\fontsize{0.33cm}{0.15cm}\selectfont
\begin{tabular}{ c | c c c c}
    \midrule
     & iRMSE & iMAE & RMSE & MAE \\
     \midrule
     stereo + CPN & {\bf 3.58} & {\bf 1.32} & {\bf 1263.19} & {\bf 343.46} \\
     \midrule
     stereo & 3.76 & 1.35 & 1281.11 & 350.28 \\
     \midrule
     monocular + CPN & {\bf 3.67} & {\bf 1.35} & {\bf 1277.24} & {\bf 349.73} \\
     \midrule
     monocular \cite{ma2018self} & 4.07 & 1.57 & 1299.85 & 350.32 \\
    \midrule
\end{tabular}
\caption{Ablation study using different supervision and prior.}
\label{tab:ablation}
\end{table}

{\small
\bibliographystyle{ieee}
\bibliography{egbib}
}

\end{document}